\documentclass[final]{cvpr}

\usepackage{times}
\usepackage{epsfig}
\usepackage{graphicx}
\usepackage{amsmath}
\usepackage{amssymb}
\usepackage{multirow}
\usepackage{booktabs}
\usepackage{color}
\usepackage{subcaption}

\usepackage[pagebackref=true,breaklinks=true,colorlinks,bookmarks=false]{hyperref}
\def\cvprPaperID{6563} 
\def\confYear{CVPR 2021}

\begin{document}

\title{Discrete-continuous Action Space Policy Gradient-based Attention for Image-Text Matching}

\author{Shiyang Yan{$^{1}$}\thanks{Corresponding Author}, \ \
Li Yu{$^1$}, \
Yuan Xie{$^{2}$}\footnotemark[1] \\
$^1$Nanjing University of Information Science and Technology, Nanjing, China \\
$^2$East China Normal University, Shanghai, China\\
{\tt\small elyotyan@gmail.com},
{\tt \small li.yu@nuist.edu.cn},
{\tt\small yxie@cs.ecnu.edu.cn} \\
}

\maketitle

\begin{abstract}
Image-text matching is an important multi-modal task with massive applications. It tries to match the image and the text with similar semantic information. Existing approaches do not explicitly transform the different modalities into a common space. Meanwhile, the attention mechanism which is widely used in image-text matching models does not have supervision. We propose a novel attention scheme which projects the image and text embedding into a common space and optimises the attention weights directly towards the evaluation metrics. The proposed attention scheme can be considered as a kind of supervised attention and requiring no additional annotations. It is trained via a novel Discrete-continuous action space policy gradient algorithm, which is more effective in modelling complex action space than previous continuous action space policy gradient. We evaluate the proposed methods on two widely-used benchmark datasets: Flickr30k and MS-COCO, outperforming the previous approaches by a large margin.
\end{abstract}

\section{Introduction}
Computer Vision and Natural Language Processing are two important areas of modern artificial intelligence, which can be processed jointly in cross-modal tasks. A large amount of research has been conducted to bridge the vision and language modalities~\cite{vinyals2015show, xu2015show, faghri2017vse++, li2019visual, lee2018stacked}. Image-text matching or retrieval is one of the critical topics in this area, which has a huge application scope in many real-world scenarios. The image-text matching requires a machine learning model to extract the high-level semantic representations and measure the similarities across modalities accordingly.
\begin{figure}
    \centering
    \includegraphics[width=\linewidth]{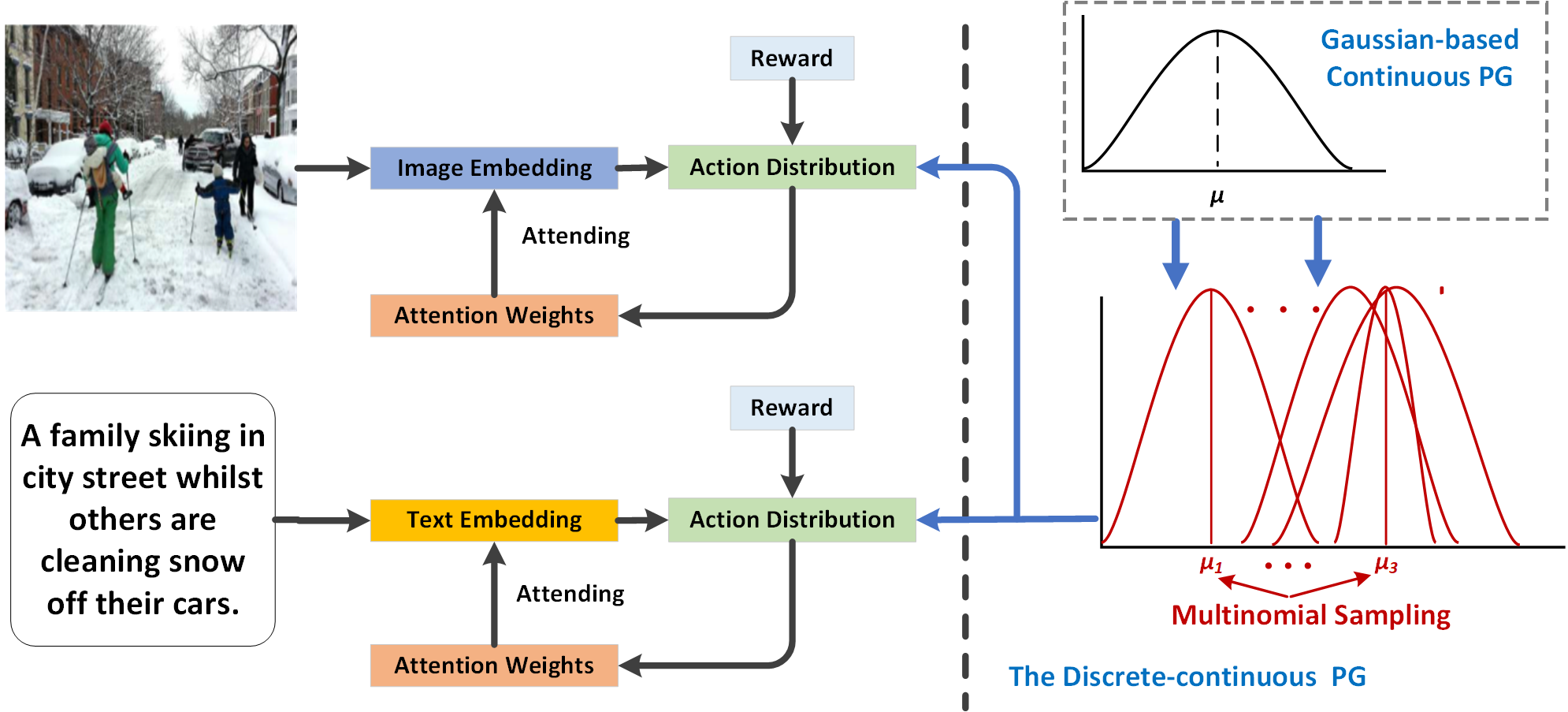}
    \caption{Our Motivations: The attention weights are utilised as a projection from each modality to a common space. Existing continuous PG assumes a simple Normal distribution. Instead, we considered the mean values as discrete actions first and then use multiple Normal distributions to form a compound distribution, which is more realistic. \textit{Best Viewed in Colour.}}
    \label{fig:illustration}
\end{figure}

Existing methods use deep learning models to extract the image and language features, and apply various metric learning techniques to automatically find the semantic similarities between the samples from the two modalities~\cite{faghri2017vse++, li2019visual, lee2018stacked}. Metric learning is powerful in visual semantic embedding as it tries to measure and manipulate the similarities between samples regardless of the domain differences. However, it is not designed to perform an explicit transformation from one modality to another, often leading to sub-optimal performance. Though there are approaches apply Instance Loss~\cite{zheng2020dual}, i.e., a classification over image and text categories, to form a multi-task learning approach with metric learning loss for image-text matching, the performance gain is limited as the Instance Loss optimises the embedding in the category domain, which does not perform explicit transformation either. An image often contains many fine-grained objects. A flat vector representation from a vanilla deep CNN model such as ResNet~\cite{he2016deep} is not powerful enough to discover these objects and their relationships. Hence, advanced methods employ image features from a pre-trained object detector~\cite{ren2015faster} and apply the visual attention mechanism~\cite{xu2015show} on these features to discriminate the important features over irrelevant ones~\cite{anderson2018bottom}. Attention mechanism plays a significant role in varying computer vision tasks. It is considered as hidden neurons in these models, but often leads to incorrect selection of image features for lacking a direct supervision~\cite{liu2016neural}.

In this paper, to make an explicit transformation and provide supervision to the attention mechanism in image-text matching, we propose a policy gradient (PG)~\cite{sutton2000policy} optimised attention adjustment over the visual and text features in image-text matching. The attention weights in our approach can be considered as a transformation from a specific modality to a common space, as the attention weights perform a vector transformation in the last image and text vectors used for matching, instead of selecting important features in the previous layers of the deep learning models~\cite{xu2015show}. The attention weights are trained by the PG method with the batch-wise ranking metrics and the instance-wise Average Precision (AP) as the reward function. These attention weights are directly optimised via PG algorithm to achieve optimal ranking results and higher AP metrics. It can be considered as a kind of supervised attention mechanism, and this supervision does not need any additional annotations. This PG-based attention mechanism is straightforward and is optimised towards the evaluation metrics. It is also more accurate than the conventional soft attention which is only a regular neuron.

To be more specific, as shown in Figure~\ref{fig:illustration}, we consider the attention weights generation as an action selecting process in PG, whose space can be flexibly pre-defined. The action space in conventional PG is discrete, which is not suitable for the feature adjustment like in the attention mechanism. One solution is applying a continuous action space PG algorithm~\cite{lillicrap2015continuous}, which consider the action space as a Gaussian distribution and sample action values from this distribution. Restricting the action distribution to Normal is not optimal, and such a hypothesis lacks theoretical and practical support. In reality, the distribution of the action space might be very complex, which cannot be described via a simple Normal distribution. Hence, we consider the action is continuous and sampled from multiple Normal distributions with a different mean ($\mu$) and standard deviation ($\sigma$) values. We first treat the $\mu$ as discrete actions, sampled from a pre-defined action space while $\sigma$ is obtained from a neural model as it is continuous. We want to use this $\mu$ and $\sigma$ to form a Normal distribution and sample continuous action from this distribution, which is applied as the attention weights to adjust the feature representations for both the visual embedding and the text embedding. Usually, in conventional PG, we do not need the $\mu$ to be trainable as we only back-propagate the gradient to the log-probabilities. In contrast, in this case, the subsequently obtained Normal distributions need the $\mu$ being able to be back-propagated to make the Normal distribution learn-able.
As there involves sampling in obtaining the $\mu$, it is not trainable in this current form. To make this $\mu$ differentiable, instead of directly using greedy sampling or $\epsilon$-greedy sampling. We use a Gumbel-softmax to relax the discreteness~\cite{jang2016categorical} and make the sampled $\mu$ trainable together with the Normal distribution. We call this method ``Discrete-continuous PG" as it involves both discrete and continuous action space, making them benefit from each other. In fact, by using the discrete and continuous action space, the action space used to sample the attention weights is a compound distribution, which can model a high complex distribution. We evaluate our algorithm and model on image-text matching task, achieving state-of-the-art performance on two widely-used benchmark datasets. To summarise, our contributions are threefold:
(1) We propose a novel attention supervision scheme for image-text matching task based on policy gradient.
(2) We propose a new Discrete-continuous policy gradient algorithm by leveraging both the discrete and continuous action space.
(3) The achieved state-of-the-art results validate the effectiveness of the attention supervisions scheme and the novel policy gradient algorithm.

\section{Related Works}
\subsection{Image-text Matching}
Frome et al.~\cite{frome2013devise} propose a feature embedding method via CNNs and Skip-Gram for cross-modal matching. They also utilise a ranking loss to measure the distance between similar pairs. Faghri et al.~\cite{faghri2017vse++} focus on the hard negative mining in the Triplet loss, with improved results. Zheng et al.~\cite{zheng2020dual} utilise an Instance Loss over a large number of categories. They find that the Instance Loss is helpful in image-text matching. Gu et al.~\cite{gu2018look} improve the cross-modal problem by looking into the generative models. Li et al.~\cite{li2019visual} propose a visual semantic reasoning framework by leveraging graph neural networks and image captioning loss. The visual semantic reasoning model can reason on the semantic relationship of the image features, with good performance.
\subsection{Attention Mechanism}
The visual attention mechanism~\cite{xu2015show} has been widely applied in many types of computer vision applications. Notably, bottom-up attention model~\cite{anderson2018bottom} is the current mainstream for image captioning, visual question answering and image-text matching. However, there is not much research on supervised attention. Gan et al.~\cite{gan2017vqs} propose a supervised attention scheme on visual question answering using attention annotations. Kamigaito et al.~\cite{kamigaito2017supervised} also use attention annotations for supervised attention in natural language processing task. Instead, we propose a supervised attention mechanism based on reinforcement learning, which can make the attention module directly optimise towards a specific goal such as AP. Also, the proposed attention module does not need any additional annotations.
\subsection{Continuous Action Space Policy Gradient}
The continuous control problem has long been investigated. For example, Lillicrap et al.~\cite{lillicrap2015continuous} propose the deep deterministic policy gradient by considering a continuous action space. Previous research has exploited the relationship between discrete and continuous action space. For instance, Dulacc-Arnold et al.~\cite{dulac2015deep} leverage the continuity in the underlying continuous action space for generalisation on discrete actions. Pazis et al.~\cite{pazis2009binary} convert the continuous control problem into discrete ones, by using a binary discrete action space. Tang et al.~\cite{tang2020discretizing} show that discretizing action space for continuous control is a simple yet powerful technique for on-policy optimisation. We also consider the combination of discrete and continuous action space for on-policy optimisation. We prove that a compound distribution is superior to a strict assumption of one Normal distribution.

\section{The Proposed Method}
\begin{figure*}
    \centering
    \includegraphics[width=0.9\linewidth]{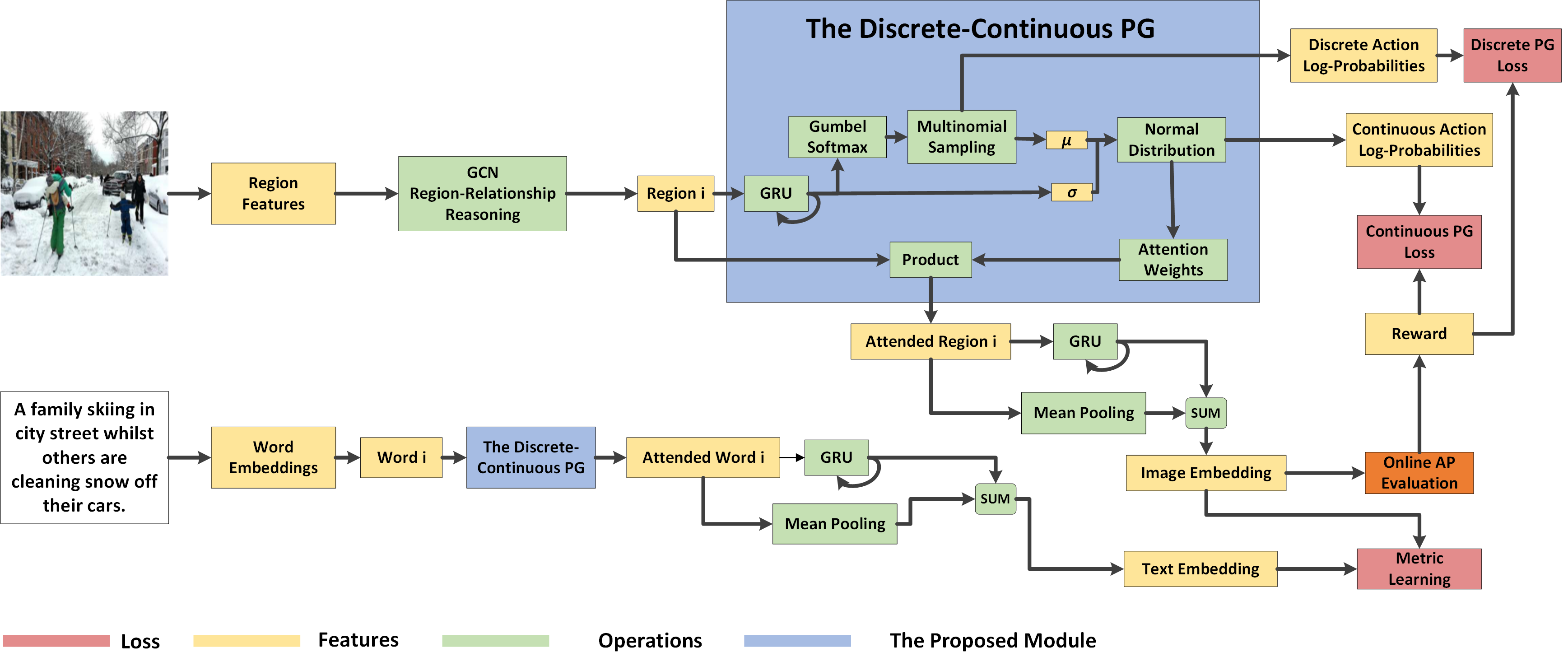}
    \caption{A schematic diagram of the proposed method: The image and text are forwarded into the model. The extracted image features are first processed via a GCN model to reason on the semantic relationships. The region features are then inputted to the proposed Discrete-continuous PG algorithm to generate attention maps, which are applied to adjust and fuse the region features subsequently. Similarly, the text embedding is also adjusted via the attention maps generated by the Discrete-continuous PG algorithm. The final image and text embedding are then connected with Metric Learning losses, the Discrete PG Loss and the Continuous PG loss for training. \textit{Best Viewed in Colour.}}
    \label{fig:method}
\end{figure*}
Our goal is to adjust the generated visual and text features to facilitate the image-text matching. We first apply Graph Convolutional Neural Networks~\cite{wu2020comprehensive} on the bottom-up attention~\cite{anderson2018bottom} features of the images, which is similar to the Visual Semantic Reasoning Networks (VSRN)~\cite{li2019visual}. Once the visual features are obtained, we then use our Discrete-continuous action space PG to generate the attention weights, which are used to adjust the visual features. Similarly, the text features are also adjusted via the proposed Discrete-continuous PG-based attention mechanism. The obtained image and text embedding are trained via multi-task loss, including the Triplet Loss, Instance Loss and Text Decoding Loss.
A schematic diagram of the proposed method is shown in Figure~\ref{fig:method}.
\subsection{Image and Text Features Extraction}
\paragraph{GCN for image region features reasoning.}
We apply a GCN model similar to VSRN approach~\cite{li2019visual}. Specifically, the semantic relationship between image region features is measured via pairwise affinity.
\begin{equation}
    Relation(F_i, F_j) = E_i(F_i)^T E_i(F_j),
    \label{gcn}
\end{equation}
where $F_i$ and $F_j$ are two bottom-up image region features obtained from Faster R-CNN detectors. $E_i$ and $E_j$ are embedding functions, which are usually matrices multiplications, which are can be learnt via backpropagation.

Then a fully-connected relationship graph $G_r = (V, E)$ is constructed. $V$ is the set of detected image region features and $E$ is the set of edges where each of the edges is described by the affinity matrices $Relation(F_i, F_j)$, which is presented in Equation~\ref{gcn}. We apply the GCN to perform reasoning on this fully-connected graph. The output of the GCN reasoning is denoted as $Image = \{I^1, ..., I^t, ...,  I^T \}$.

\paragraph{Text Embedding.}
Given one-hot text representations, represented as $w$, a linear word embedding layer is constructed to obtain the word representations, represented as $W_e = \{w_e^1, ..., w_e^i, ..., w_e^N\}$,  where $w_e^i = word\text{-}embedding(w^i)$.

\subsection{The Proposed Discrete-continuous Action Space PG}
PG is usually with discrete action space for two reasons: many control problems are modeled in discrete action space which leads to high performance as it can model complex action distribution. However, when meeting with continuous action space control problem, we have to develop corresponding PG algorithms. However, as discussed previously, continuous action space PG normally assumes the actions follow a Normal distribution, which is too strict. We propose an approach to essentially sample the continuous action from a compound distribution, which can better model the real distribution.
\paragraph{Discrete Action Sampling.}
As shown in Figure~\ref{fig:method}, we first model the attention weights generation process as a finite Markov Decision Process (MDP) and sample a discrete action by using Multinomial Sampling. We define $n$ action categories, i.e., $A=\{a_1, a_2, ..., a_n\}$, The state space contains the input region features and the attention weights generated so far, which are $s_t = \{ I^0, Att^{0}..., I^{t-1}, Att^{t-1}\}$. The policy is parametrised via a GRU model to explore the environment and sample the action. More formally:
\begin{equation}
\begin{split}
   & h^t = GRU_{mdp}(I^t, h^{t-1}), \ \  t = 1,..., T, \\
   &  F_{I}^t = h^t, \\
   & a^{t} = F_{I}^t*W_{\mu}^t, \\
   & a_{g}^t = Gumbel\text{-}softmax(a^{t}), \\
     &  a_{s}^t\sim Multinomial(a_{g}^t), \\
   & logprob_{a}^t = \log a_{g}^t(a_{s}^t),  \\
\end{split}\label{mu}
\end{equation}
where $I^t$ is the $t_{th}$ image feature after the GCN reasoning. $GRU_{mdp}$ is the Gated Recurrent Unit (GRU) used to model the attention weights generation problem as MDP. $W_{\mu}^t \in \mathcal{R}_{s \times n}$ are the weights need to be learnt. $s$ is the size of the feature vector. $a_{g}^t$ is the probability of each actions after the $Gumbel\text{-}softmax$ activation.
\begin{equation}
\begin{split}
  & \mu^t = Logistic \left(\frac{a_s^t}{n}\right),    \\
  & {std}^t =  F_{I}^t*W_{std}^t, \\
\end{split}
\end{equation}
where $W_{std} \in \mathcal{R}_{s \times 1}$ are the weights need to be learnt.

\paragraph{Continuous Action Sampling.}
The sampled $\mu$ and $\sigma$ form a Normal Distribution, described as follows:
\begin{equation}
\begin{split}
&     Sample \sim \mathcal{N}(\mu^t, \sigma^t), \\
&     Att^t = Sigmoid(Sample),
\end{split}
\end{equation}
where $Att^t$ are the attention weights sampled from this particular Normal Distribution. The log probabilities of this Normal Distribution is expressed as:
\begin{equation}
\begin{split}
 &   \log{\left(f\left(Att^t;\,\mu^t,{\sigma^t}^2\right)\right)}
=   \\
& - \frac{n}{2} \log{\left(2 \pi\right)}
- \frac{n}{2} \log{\left({(\sigma^t)}^2\right)} \\
& - \frac{1}{2 {(std^t)}^2} \sum{{\left(Att - \mu^t \right)}^2}.
\end{split}\label{log_cont}
\end{equation}

\paragraph{Discrete PG Optimisation.}
To be simple and efficient, we formulate the PG as an on-line learning method, specifically, the REINFORCE algorithm~\cite{williams1992simple}. The PG for discrete action space is then to maximise the long-term reward with the following expression:
\begin{equation}
\begin{split}
&    \nabla_\theta J(\theta) =  \\
& \mathbb{E}_{\tau \sim \pi_\theta(\tau)}
\left[ \left(\sum_{t=0}^{T} \nabla_\theta \log{\pi_\theta}(a_t \mid s_t)\right) \left(\sum_{t=0}^{T} r(s_t, a_t)\right)\right],
\end{split}\label{pgmu}
\end{equation}
we use the one sample Monte-Carlo to approximate the accumulative reward, i.e., $\sum_{t=0}^{T} r(s_t, a_t) = \sum_{t=0}^{T} \mathcal{R}$, where $\mathcal{R}$ is the reward and will be defined later. Also, $\log{\pi_\theta}(a_t \mid s_t) = logprob_{a}^t$, which is obtained from Equation~\ref{mu}. Hence, Equation~\ref{pgmu} can lead to a PG loss function as follows,

\begin{equation}
\begin{split}
&    Loss_{PG_{D}} = - \sum_{b=1}^{B}
\left[ \left(\sum_{t=0}^{T} \nabla_\theta logprob_{a}^t \right) \left(\sum_{t=0}^{T} \mathcal{R} \right)\right],
\end{split}\label{loss_discrete}
\end{equation}
where $B$ is the size of each mini-batch. Note that the negative notation on the right-hand side means that we want to minimise the loss so as to maximise the $\mathcal{R}$.

\paragraph{Continuous PG Optimisation.}
Equation~\ref{log_cont} provides a straightforward definition of the log probabilities of a Normal Distribution. Similarly, the PG loss for the continuous action space is presented as follows:
\begin{equation}
\begin{split}
&    Loss_{PG_{C}} = \\
& - \sum_{b=1}^{B}
\left[ \left(\sum_{t=0}^{T} \log{\left(f\left(Att^t;\,\mu^t,{\sigma^t}^2\right)\right)} \right) \left(\sum_{t=0}^{T} \mathcal{R}\right) \right].
\end{split}\label{loss_cont}
\end{equation}

\paragraph{Reward Function Formulation.}
The reward signal is of vital significance as it guides the attention generation process, which is the initial goal of PG method. The reward signal is obtained from an on-line evaluation of the image and text embedding using R@K and Average Precision (AP). We consider a batch of samples as the gallery, and each sample as a query to compute the instance-wise AP.
Specifically, we treat each of the samples as one category, and calculate the R@1 and AP of it on-line in a batch of samples. The reward signal can thus be expressed as a linear combination of the R@1 and the AP results:
\begin{equation}
\mathcal{R} = R\text{@}1 + AP,
\end{equation}
we then use this reward to guide the proposed PG algorithm to generate attention weights to automatically adjust the image and text features to formulate a more effective embedding for the image-text matching task. To further reduce the variance and make the PG training more stable, we additionally apply a PG baseline, which is an average of the rewards from all the other instances within a batch, expressed as:
 \begin{equation}
 b_k = \frac{1}{K-1}\sum_{j\neq k}\mathcal{R}_j,
 \end{equation}
where $K$ is the batch size, $b_k$ is the baseline for $k_{th}$ instance and $R_j$ is the reward of $j_{th}$ instance. We apply a coefficient $\beta = 0.5$ over the baseline, which is empirically better.

\subsection{Feature Fusion}
The image embedding can be adjusted by using the generated attention weights. Recall the image region features as $Image = \{I^1, ..., I^t, ..., I^T\}$, and the generated attention weights are $ATT = \{Att^1, ..., Att^t, ..., Att^T\}$, we use element-wise multiplication to adjust the image region features with the attention weights.
\begin{equation}
\begin{split}
& I_{A} = Image*(\lambda * ATT^{I}), \\
& h_g^t = GRU_{gr}^{I}(I_{A}^t,  h_{g}^{t-1}), \ t = 1, ..., T,   \\
& I_{E} = h_g^T + [\sum_{t=1}^T I_{A}]/T,
\end{split}\label{fusion_i}
\end{equation}
where $I_{A}$ stands for adjusted image region features. $GRU_{gr}^{I}$ is used to perform global reasoning of the adjusted image features. The fused features involve a summation of the outputs of the $GRU_{gr}$ and the adjusted image region features. $I_{E}$ is the image embedding.

Similarly, we apply the same approach to the text embedding generation. Note that we directly apply the proposed Discrete-continuous PG on the word embedding $W_e$.

Then the feature adjustment and fusion of text embedding generation can be presented as follows:

\begin{equation}
\begin{split}
& T_{A} = W_e*(\lambda * ATT^{T}),\\
& h_g^i = GRU_{gr}^{T} (T_{A}^i,  h_{g}^{i-1}), \ i = 1, ..., N,   \\
& T_E = h_g^N + [\sum_{i=1}^N T_{A}]/N,
\end{split}\label{fusion_t}
\end{equation}
where $T_{A}$ is the adjusted text features and $ATT^{T}$ are the attention weights generated for text embedding. $T_{E}$ is the text embedding.
\subsection{Loss Functions}
To fulfill the image-text matching task, we apply cross-modal Triplet Loss, Instance Loss, Text Decoding Loss, and together with the proposed PG loss, to train the model. The final loss objective function of the model is described as follows:
\begin{equation}
\begin{split}
    \mathbb{L} = & Loss_{triplet} + Loss_{xe} + loss_{td}^{I} + loss_{td}^{T} \\
    & + Loss_{PG_c}^{I} + Loss_{PG_d}^{I} + Loss_{PG_c}^{T} + Loss_{PG_d}^{T},
\end{split}
\end{equation}
where $Loss_{triplet}$ is the hinge-based Triplet ranking loss~\cite{faghri2017vse++, karpathy2015deep, lee2018stacked}. The $Loss_{xe}$ is the cross-entropy classification loss which treats each instance as one class categories~\cite{zheng2020dual}. The $Loss_{td}^{I}$ and $Loss_{td}^{T}$ are the Image-to-Text Decoding Loss and Text-to-Text Decoding Loss, respectively. They decode the image or text embedding into sentences. Note the weights of the Text Decoding Module are shared between image and text branches.

The Triplet loss is expressed as follows:
\begin{equation}
\begin{split}
   Loss_{metric} = & [\alpha - S(I, T) + S(I, \hat{T})]_{+}  + \\
   & [\alpha - S(I, T) + S(\hat{I}, T)]_{+},
\end{split}
\end{equation}
where $\alpha$ is the margin hyper-parameter.$[x]_{+} = max(x, 0)$. $S(\cdot)$ is the similarity function. $\hat{I}$ and $\hat{T}$ are the hardest negatives for a positive pair $(I, T)$.

For the Text Decoding Loss, We apply the convolutional image captioning model~\cite{AnejaConvImgCap17} as the decoder of the image and text decoding module. We use the same loss functions as in~\cite{AnejaConvImgCap17}, which has a parallel training capability for text decoding and is much efficient than the RNN-based one.

\begin{table}[!t]
	\centering
	\resizebox{\linewidth}{!}{
	\begin{tabular}{llllllll}
		\toprule
		\multirow{2}{*}{Networks} & \multirow{2}{*}{Methods} & \multicolumn{3}{c}{Caption Retrieval} & \multicolumn{3}{c}{Image Retrieval} \\
		\cline{3-8}
		& & R@1 & R@5 & R@10  & R@1 & R@5 &  R@10  \\
		\midrule
		\multirow{2}{*}{AlexNet} &
		DSVA~\cite{karpathy2015deep} & 22.2 & 48.2 & 61.4 & 15.2 & 37.7 & 50.5 \\
		& HMlstm~\cite{niu2017hierarchical} & 38.1 & - & 76.5 & 27.7 & - & 68.8 \\
		\midrule
		\multirow{4}{*}{VGG} &
		FV~\cite{klein2015associating} & 35.0 & 62.0 & 73.8 & 25.0 & 52.7 & 66.0 \\
		& VQA~\cite{lin2016leveraging} & 33.9 & 62.5 & 73.8 & 25.0 & 52.7 & 66.0 \\
		& SMlstm~\cite{huang2017instance} & 42.5 & 71.9 & 81.5 & 30.2 & 60.4 & 72.3 \\
		& 2wayN~\cite{eisenschtat2017linking} & 49.8 & 67.5 & - & 36.0 & 55.6 & - \\
		\midrule
		\multirow{3}{*}{ResNet} &
		RRF~\cite{liu2017learning} & 47.6 & 77.4 & 87.1 & 35.4 & 68.3 & 79.9 \\
		& VSE~\cite{faghri2017vse++} & 52.9 & 79.1 & 87.2 & 39.6 & 69.6 & 79.5 \\
		& SCO~\cite{huang2018learning} & 55.5 & 82.0 & 89.3 & 41.1 & 70.5 & 80.1 \\
		\midrule
		\multirow{2}{*}{Faster R-CNN}  &
	    SCAN~\cite{lee2018stacked} & 67.4 & 90.3 & 95.8 & 48.6 & 77.7 & 85.2 \\
	    &	VSRN~\cite{li2019visual}  & 71.3 & 90.6 & {96.0} & 54.7 & 81.8 & 88.2 \\
			\cline{2-8}
		&	\textbf{Ours} & \textbf{82.8} & \textbf{95.9} & \textbf{97.9} & \textbf{62.2} & \textbf{89.3} & \textbf{93.8} \\
	\bottomrule
	\end{tabular}
	}	\caption{Comparison of the image-text matching on Flickr30k Dataset.}\label{table:fiker}
\end{table}

\begin{table}[!t]
	\centering
	\resizebox{\linewidth}{!}{
	\begin{tabular}{llllllll}
		\toprule
		\multirow{2}{*}{Networks} & \multirow{2}{*}{Methods} & \multicolumn{3}{c}{Caption Retrieval} & \multicolumn{3}{c}{Image Retrieval} \\
		\cline{3-8}
		& & R@1 & R@5 & R@10  & R@1 & R@5 &  R@10  \\
		\midrule
		\multirow{2}{*}{AlexNet} &
		DSVA~\cite{karpathy2015deep} & 38.4 & 69.9 & 80.5 & 27.4 & 60.2 & 74.8 \\
		& HMlstm~\cite{niu2017hierarchical} & 43.9 & - & 87.8 & 36.1 & - & 86.7 \\
		\midrule
		\multirow{4}{*}{VGG} &
		FV~\cite{klein2015associating} & 39.4 & 67.9 & 80.9 & 25.1 & 59.8 & 76.6 \\
		& VQA~\cite{lin2016leveraging} & 50.5 & 80.1 & 89.7 & 37.0 & 70.9 & 82.9 \\
		& SMlstm~\cite{huang2017instance} & 53.2 & 83.1 & 91.5 & 40.7 & 75.8 & 87.4 \\
		& 2wayN~\cite{eisenschtat2017linking} & 55.8 & 75.2 & - & 39.7 & 63.3 & - \\
		\midrule
		\multirow{3}{*}{ResNet} &
		RRF~\cite{liu2017learning} & 56.4 & 85.3 & 91.5 & 43.9 & 78.1 & 88.6 \\
		& VSE~\cite{faghri2017vse++} & 64.6 & 89.1 & 95.7 & 52.0 & 83.1 & 92.0 \\
		& GXN~\cite{gu2018look} & 68.5 & - & 97.9 & 56.6 & - & 94.5 \\
		& SCO~\cite{huang2018learning} & 69.9 & 92.9 & 97.5 & 56.7 & 87.5 & 94.3 \\
		\midrule
		\multirow{3}{*}{Faster R-CNN}
		& SCAN~\cite{lee2018stacked} & 72.7 & 94.8 & 98.4 & 58.8 & 88.4 & 94.8 \\
		& VSRN~\cite{li2019visual}  & 76.2 & 94.8 & \textbf{98.2} & 62.8 & \textbf{89.7} & 95.1 \\
		\cline{2-8}
		& \textbf{Ours} & \textbf{84.0} & \textbf{95.8} & {97.8} & \textbf{63.9}  & {88.9} & \textbf{95.6} \\
		\bottomrule
	\end{tabular}
	}	\caption{Comparison of the image-text matching on MS-COCO Dataset of 1K test set.}\label{table:coco1k}
\end{table}

\begin{table}[!t]
	\centering
	\resizebox{\linewidth}{!}{
	\begin{tabular}{llllllll}
		\toprule
		\multirow{2}{*}{Networks} & \multirow{2}{*}{Methods} & \multicolumn{3}{c}{Caption Retrieval} & \multicolumn{3}{c}{Image Retrieval} \\
		\cline{3-8}
		& & R@1 & R@5 & R@10  & R@1 & R@5 &  R@10  \\
		\midrule
		\multirow{2}{*}{AlexNet} &
		DSVA~\cite{karpathy2015deep} & 11.8 & 32.5 & 45.4 & 8.9 & 24.9 & 36.3 \\
		\multirow{4}{*}{VGG} &
		FV~\cite{klein2015associating} & 17.3 & 39.0 & 50.2 & 10.8 & 28.3 & 40.1 \\
		\hline
		& VQA~\cite{lin2016leveraging} & 23.5 & 50.7 & 63.6 & 16.7 & 40.5 & 53.8 \\
		& OEM~\cite{huang2017instance} & 23.3 & - & 84.7 & 31.7 & - & 74.6 \\
		\midrule
		\multirow{3}{*}{ResNet}
		& VSE~\cite{faghri2017vse++} & 41.3 & 69.2 & 81.2 & 30.3 &  59.1 & 72.4 \\
		& GXN~\cite{gu2018look} & 42.0 & - & 84.7 & 31.7 & - & 74.6 \\
		& SCO~\cite{huang2018learning} & 42.8 & 72.3 & 83.0 & 33.1 & 62.9 & 75.5 \\
		\midrule
		\multirow{2}{*}{Faster R-CNN}
		& SCAN~\cite{lee2018stacked} & 50.4 & 82.2 & 90.0 & 38.6 & 69.3 & 80.4 \\
		& VSRN~\cite{li2019visual} & 53.0 & 81.1 & 89.4 & 40.5 & 70.6 & 81.1 \\
		\cline{2-8}
		& \textbf{Ours}  & \textbf{68.7} & \textbf{88.7} & \textbf{93.0} & \textbf{46.2}  & \textbf{77.8} & \textbf{85.5}   \\
		\bottomrule
	\end{tabular}
	}	\caption{Comparison of the image-text matching on MS-COCO Dataset of 5K test set.}\label{table:coco5k}
\end{table}

\begin{table}[!t]
	\centering
	\resizebox{\linewidth}{!}{
	\begin{tabular}{llllllll}
		\toprule
		\multirow{2}{*}{Methods} & \multicolumn{3}{c}{Caption Retrieval} & \multicolumn{3}{c}{Image Retrieval} \\
		\cline{2-7}
		& R@1 & R@5 & R@10  & R@1 & R@5 &  R@10  \\
		\midrule
		Triplet Loss & 68.2 & 88.8 & 93.6 & 52.6 & 75.5 & 86.0 \\
		Triplet + Instance & 69.3 & 89.5 & 93.5 & 52.1 &  77.8 & 87.8  \\
		Triplet + Instance + Text (\textbf{Baseline}) & 70.9 &  89.0 & 93.5 & 52.2 & 78.1 & 87.2\\
		\midrule
		{Baseline} + Discrete PG & 78.0 &93.4 & 94.6 & 56.0 & 80.6 & 89.1 \\
        Continuous PG & 76.8 & 90.2 & 91.6 & 54.3 & 78.4 & 87.4 \\
        Multi-head Continuous PG & 78.1 & 91.6 & 92.0 & 56.2 & 79.9 & 89.2\\
		{Baseline} + Our PG (\textbf{Our scheme})  & {81.0} & {94.5} & {97.1} & {60.6} & {86.5} & {92.4} \\
		\midrule
		{Our Scheme} + Reward R@1 & 79.3 & 94.5 & 96.0 & 57.8 & 82.3 & 90.2 \\
		{Our Scheme} + Reward AP & 80.4 & 95.2 & 96.8 & {60.8} &  84.9 & {92.7} \\
		{Our Scheme} + Reward R@1+AP & {81.0} & {94.5} & {97.1} & 60.6 & {86.5} & 92.4 \\
		\midrule
		{Our scheme} + ($\lambda = 10$)  & {80.2} & {95.3} & {96.9} & {58.4} & {82.2} & {90.6} \\
		{Our scheme} + ($\lambda = 20$) & {81.0} & {94.5} & {97.1} & {60.6} & {86.5} & {92.4} \\
		{Our scheme} + ($\lambda = 30$) & 79.2 & 94.7 & 96.9 & 57.1 & 83.0 & 90.3\\
\midrule
        {Our scheme} (+PG baseline) & {80.3} &94.6 &97.5 &60.8 &86.9 & 92.4 \\
        {Our scheme} (+multi-head) &  81.4 & {94.9} & {97.7} & {61.2} & {87.5} & {92.6}\\
        \textbf{Our scheme} \textbf{(+GloVe Embedding)} & \textbf{82.8} & \textbf{95.9} & \textbf{97.9} & \textbf{62.2} & \textbf{89.3} & \textbf{93.8} \\
        \bottomrule
	\end{tabular}
	}	
	\caption{Ablation Studies on Fickr30k Dataset.}\label{table:ablation}
\end{table}
\begin{figure*}
\centering
\includegraphics[width=0.9\textwidth]{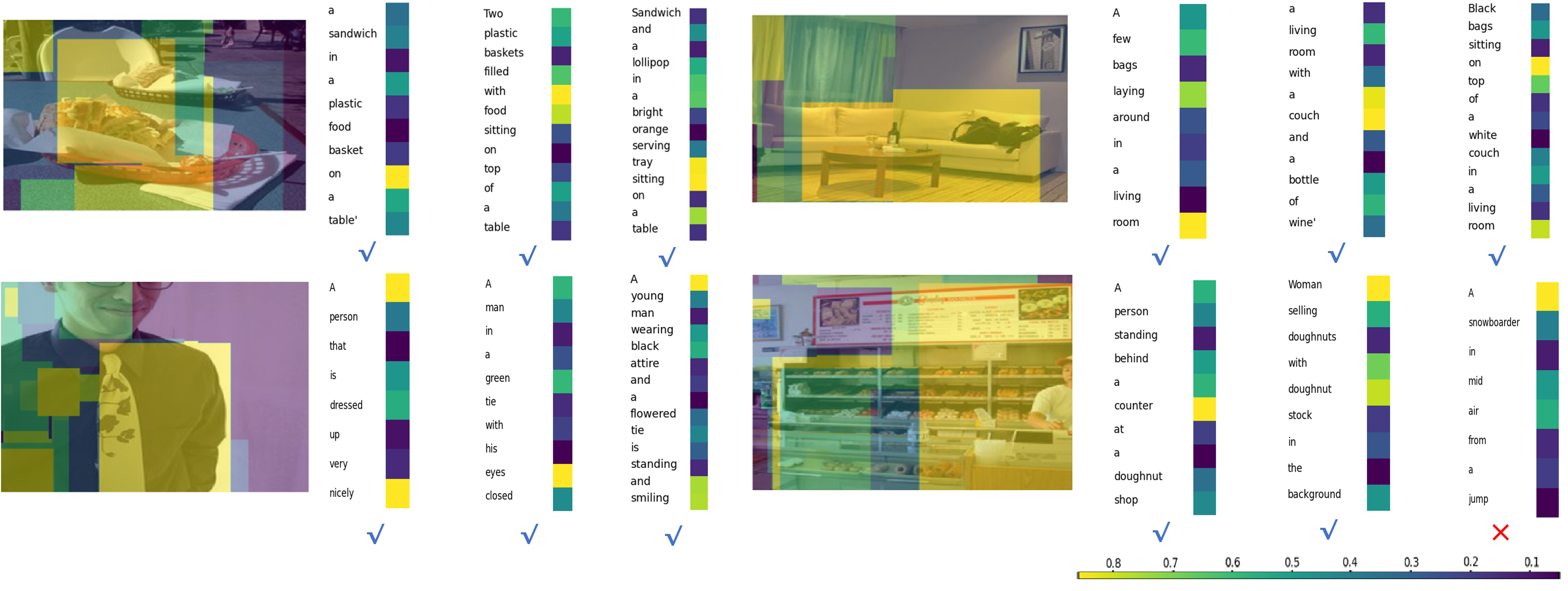}
\caption{Visualisation of the caption retrieval results and attention mechanism.  We select top 3 retrieval results where a \checkmark means the retrieval is correct whilst the $\times$ indicates a wrong result. \textit{Best Viewed in Colour.}}
\label{img:vis1}
\end{figure*}

\begin{figure*}
\centering
\includegraphics[width=0.9\textwidth]{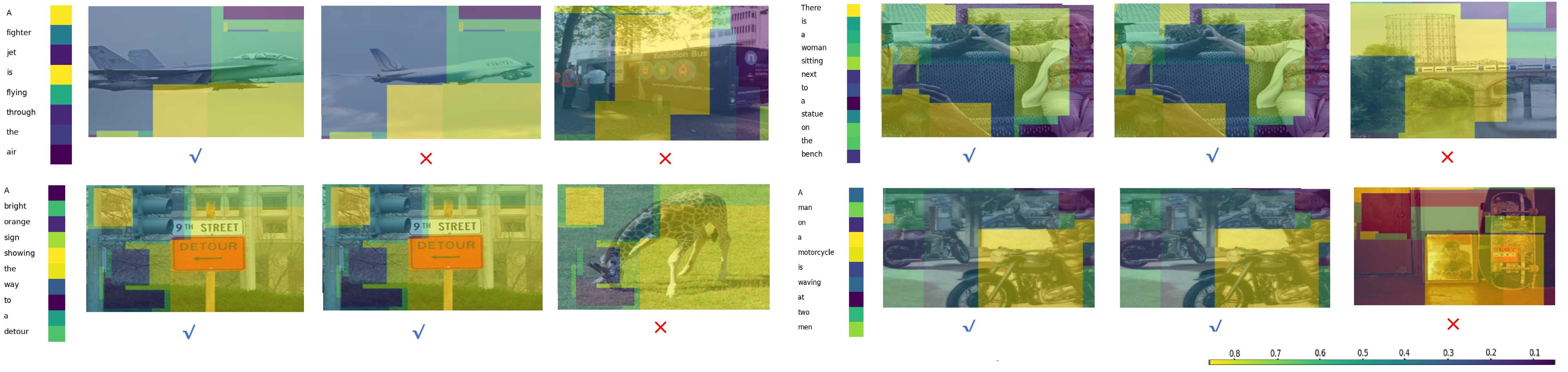}
\caption{Visualisation of the image retrieval results and attention mechanism. \textit{Best Viewed in Colour.}}
\label{img:vis2}
\end{figure*}

\begin{figure*}
		\centering
		\begin{subfigure}[t]{0.245\linewidth}
			\centering
			\includegraphics[height = 2.5cm, width=\textwidth]{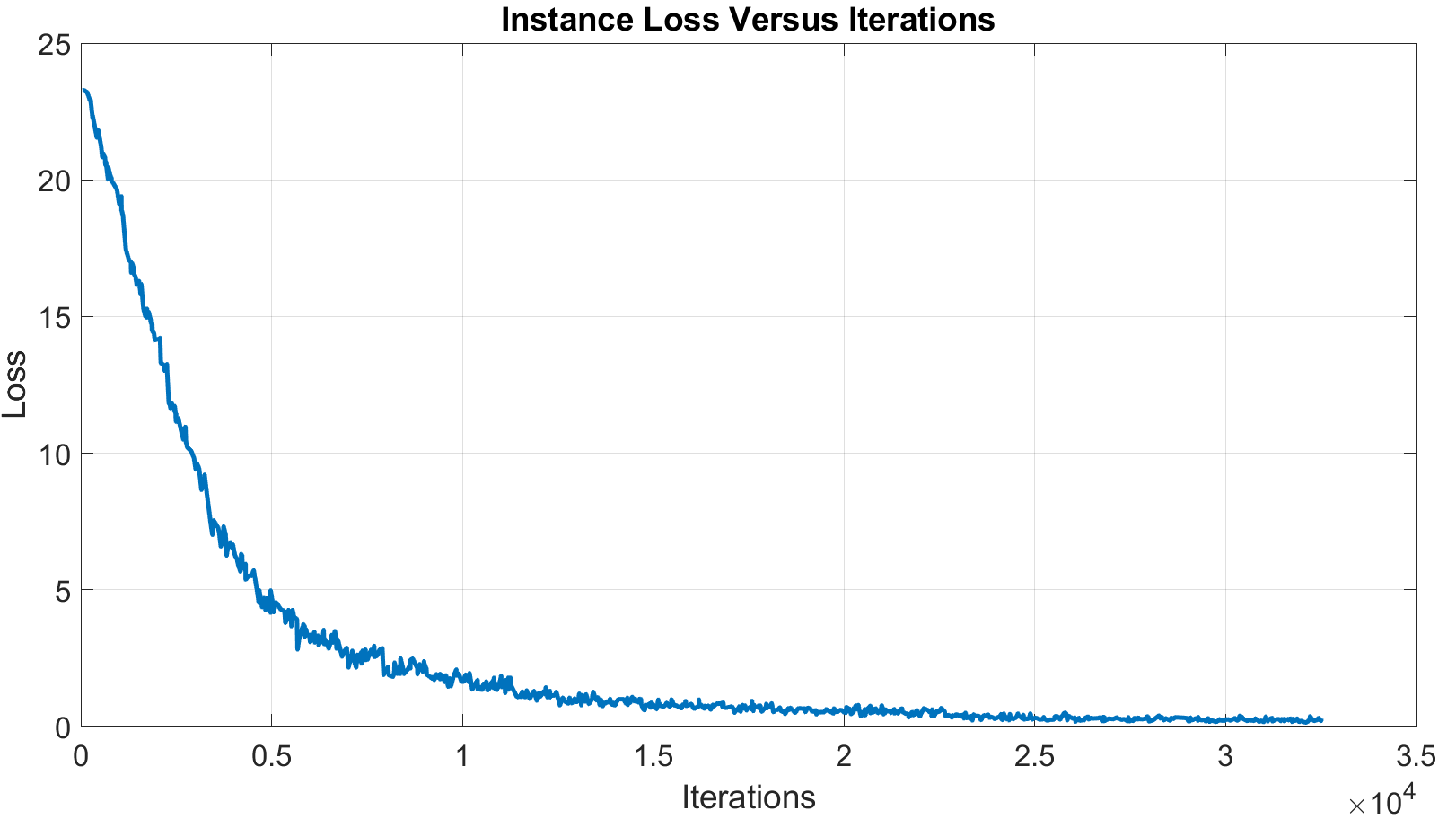}
		\end{subfigure}
		\hfill
		\begin{subfigure}[t]{0.245\linewidth}
			\centering
			\includegraphics[height = 2.5cm, width=\textwidth]{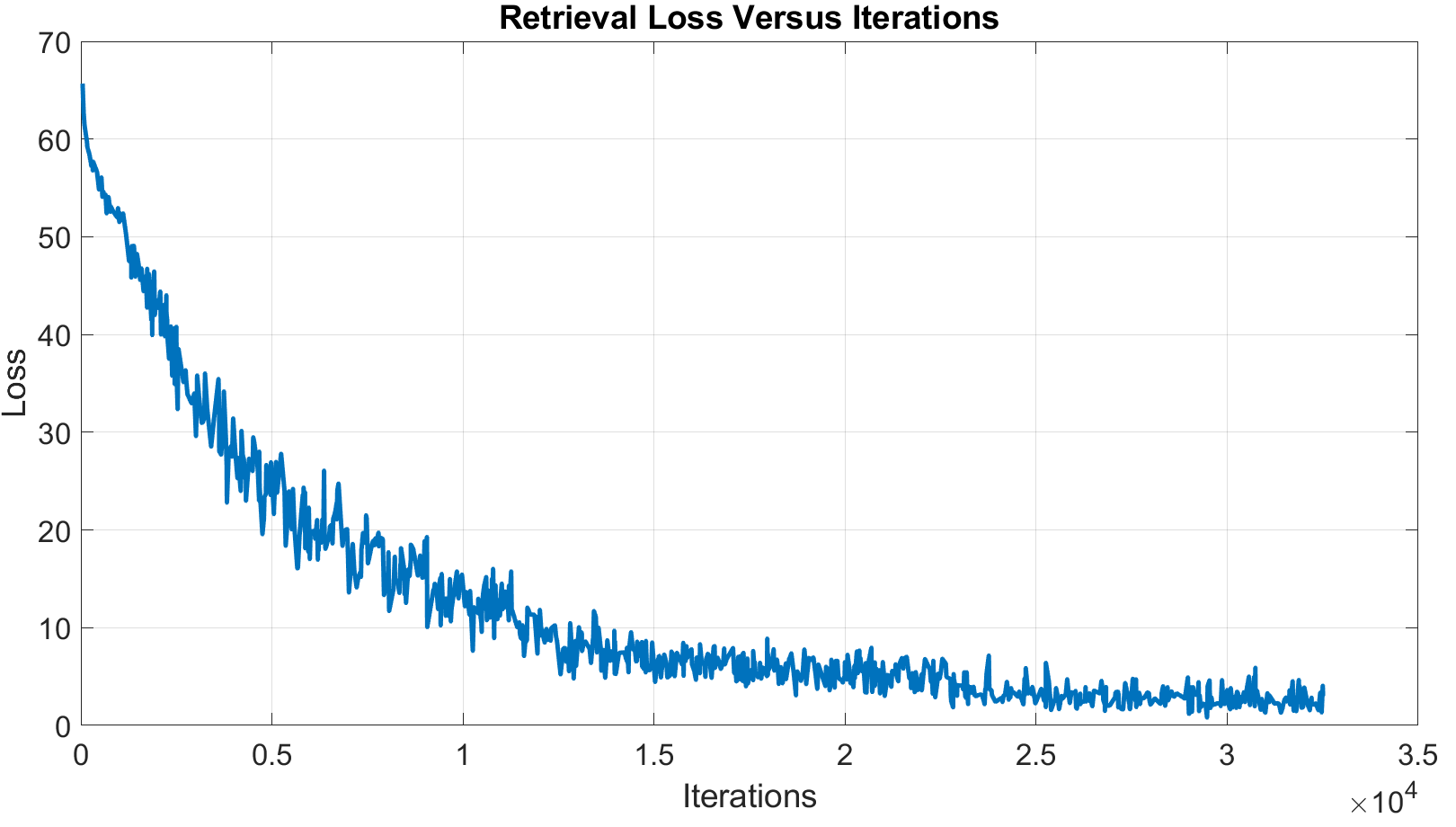}
		\end{subfigure}
		\begin{subfigure}[t]{0.245\linewidth}
			\centering
			\includegraphics[height =2.5cm,width=\textwidth]{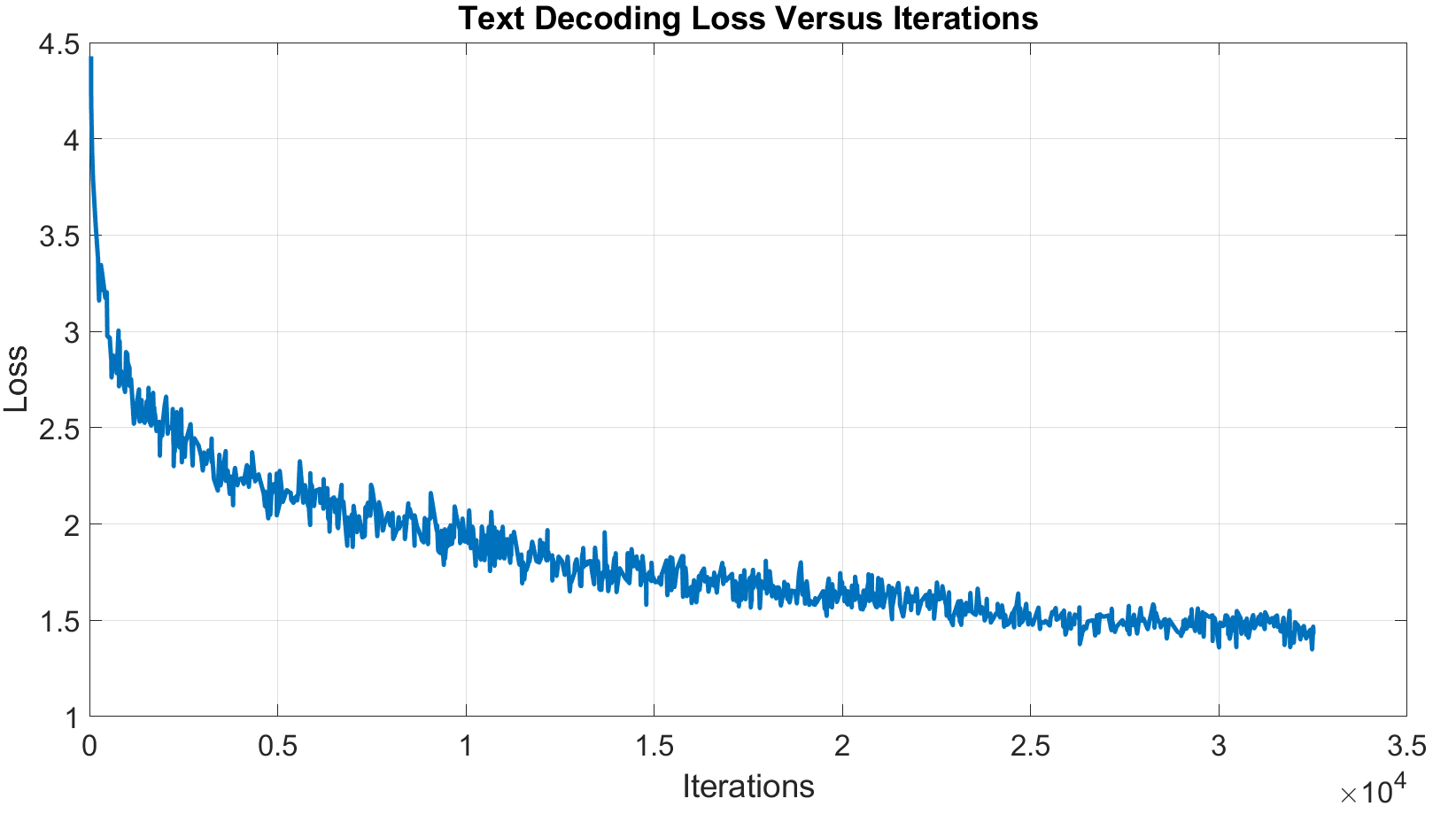}
          \end{subfigure}
        \begin{subfigure}[t]{0.245\linewidth}
			\centering
			\includegraphics[height = 2.5cm,width=\textwidth]{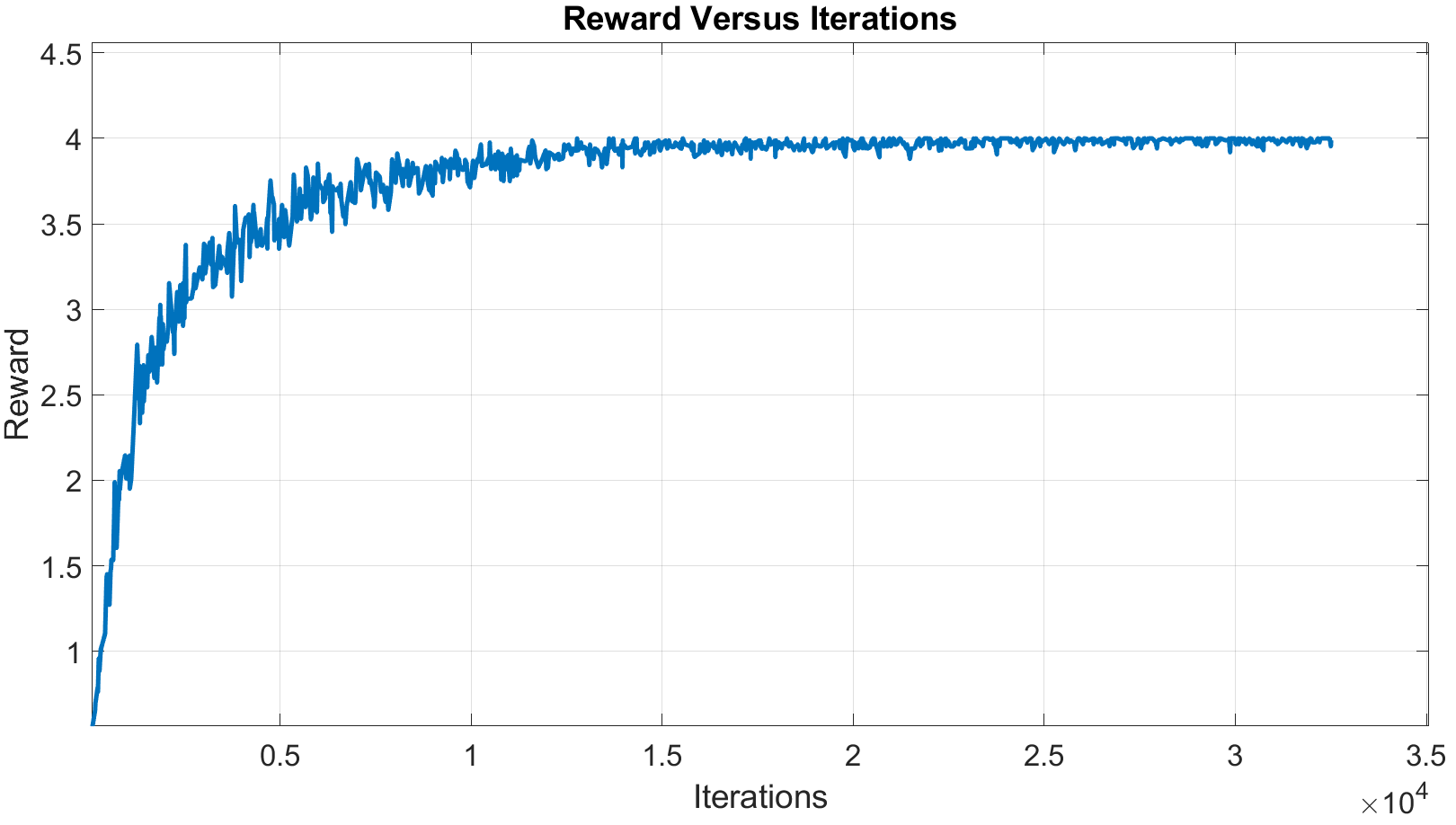}
          \end{subfigure}
		\caption{The Instance Loss, Triplet (Retrieval) Loss, Text Decoding Loss and Reward curves are shown in the figures.}
		\label{img:curves}
	\end{figure*}

\section{Experiments}
To evaluate the effectiveness of the proposed Discrete-continuous PG algorithm, we follow previous research and perform two kinds of experiments which include sentence retrieval using image and image retrieval using a sentence.
\subsection{Datasets and Protocols}
We evaluate the performance of our method on the Flickr30K~\cite{young2014image} and Microsoft-COCO datasets~\cite{lin2014microsoft}. Flickr30K contains 31,783 images. Each image corresponds to 5 human-annotated text descriptions. We use the standard training, validation and testing split ~\cite{karpathy2015deep}, which consist of 28,000 images, 1, 000 images and 1, 000 images, respectively.  We follow the splits of~\cite{faghri2017vse++, gu2018look, karpathy2015deep, lee2018stacked} for MS-COCO dataset, which includes 113, 287 images for training,  5, 000 images for validation and 5, 000 images for testing. Each image has five captions. We use the same evaluation protocol as the previous research~\cite{faghri2017vse++, gu2018look, lee2018stacked, li2019visual}, which is the recall performance at K (R@K) defined as the proportion of queries for which the correct item is retrieved in the nearest K samples to each query.

\subsection{Implementation Details}
We build our model based on PyTorch~\cite{paszke2017automatic}. We use the pre-trained bottom-up attention image features provided by~\cite{li2019visual}. The word embedding size is 300 and the dimension of the image and text embedding is 2048. The hidden size of the GRU modules used in our model is 2048. We pre-define 100 discrete action categories which are $\{0, 1, 2, ..., a_i, ... 100\}$ , where $a_i$ corresponds an action of enlarging the features with a value of $a_i/\lambda$, where $\lambda$ is a hyper-parameter. Note the choice of the number of action categories is mostly empirical. We choose 100 as it is close to the maximum number of regions of an image, and also close to the maximum number of words of a sentence, which is powerful enough to describe the difference between each item of the image regions and the sentence. The detailed explanation is presented in Equation~\ref{fusion_i} and~\ref{fusion_t}. For training, we apply Adam optimiser~\cite{kingma2014adam} to train the model with 30 epochs with a mini-batch size of 128. We start the training with a learning rate of 4e-4 for 15 epochs and lower the learning rate to 4e-5 for another 15 epochs. We apply the early stopping tricks to select the model which performs the best in the validation set. For the cross-modal Triplet ranking loss, the margin is set 0.2 for all the experiments. For the classification loss, there are 29, 783 categories for Flickr30K dataset and 113, 287 categories for MS-COCO dataset. We perform all the experiments on a server equipped with an Nvidia Geforce 2080-TI GPU card, and a Windows 10 operating system.

\subsection{Comparison with the State-of-the-art}
\paragraph{Results on Flickr30k.}
We show the results on the Flickr30k dataset and comparison with the current state-of-the-art methods in Table~\ref{table:fiker}. We also indicate the backbone networks that used for each of the state-of-the-art methods, such as AlexNet~\cite{krizhevsky2017imagenet}, VGG~\cite{simonyan2014very}, ResNet~\cite{he2016deep}, Faster R-CNN~\cite{ren2015faster}.
The proposed method outperforms other approaches by a large margin. SCAN~\cite{lee2018stacked} and VSRN~\cite{li2019visual} are two approaches that close to ours. Our method is different from them mainly on the proposed PG-based supervised feature attention mechanism as both VSRN and our method use the same cross-modal Triplet loss and the Text Decoding Loss. Hence, the main performance gain is from the proposed Discrete-continuous PG algorithm, which is effective in improving the existing baseline model that is similar to the VSRN model~\cite{li2019visual}. Specifically, we achieve 82.8\% R@1 in captioning retrieval using the image, and 62.2\% R@1 image retrieval using the caption.

\paragraph{Results on MS-COCO.}
We present the experimental results on the 1K and 5K MS-COCO dataset and comparison with the state-of-the-art models in Table~\ref{table:coco1k} and Table~\ref{table:coco5k}, respectively. For the 1K testing protocol, the results are obtained by averaging over 5 folds of 1K test images. When comparing with the current best method SCAN~\cite{lee2018stacked} and VSRN~\cite{li2019visual}, we follow the same strategy to combine results from two trained proposed models by averaging their predicted similarity scores. As shown in Table~\ref{table:coco1k}, our proposed model achieves 84.0\% R@1 on caption retrieval using an image, and 63.9 \% R@1 on image retrieval using the caption, respectively.  The results outperform the VSRN and SCAN by a large margin. For the 5K testing protocol, we evaluate the proposed model by using the whole 5K testing samples. From Table~\ref{table:coco5k}, it is obvious that our method achieves the new state-of-the-art, with 68.7\% R@1 and on 46.2\% R@1 on caption retrieval using image and image retrieval using the caption, respectively.

\subsection{Ablation Studies}
\paragraph{Baseline.}
We perform ablation studies on each component of the proposed model. which are shown in Table~\ref{table:ablation}. We first evaluate the model with only Triplet Loss, with relatively poor results. Adding an Instance Loss to the model brings an limited increase in the ranking results. Similarly, the Text Decoding Loss also improves the performance of the model, which proves that it is helpful to narrow the domain gap between different modalities. Our baseline model include all of the three Loss functions.
\paragraph{The Impact of the Discrete-continuous PG Method.}
Based on the baseline model, to validate the superiority of the proposed Discrete-continuous action space policy gradient algorithm, we first compare it with the conventional discrete action space policy gradient scheme. To realise the Discrete PG scheme, we remove the continuous action space sampling and utilise the discrete action directly as the attention weights. The proposed method yields better results than the Discrete PG scheme. Second, we solely apply a single Gaussian-based continuous action space PG scheme. The results of our scheme is also better than the single Gaussian PG as we form a complex distribution which better describe the real distribution of the action space, the results are shown in Table~\ref{table:ablation}.
\paragraph{The Impact of Different Reward Function.}
We then perform ablation studies on the reward function, the results show that using the batch-wise R@1 combined the instance-wise AP as the reward has the best performance. Note that AP alone is better than R@1 reward, as the AP evaluation is more comprehensive and instance-wise reward is more accurate than the batch-wise one. To further reduce the variance and make the PG training more stable, we additionally apply a PG baseline. The impact of the PG baseline is evaluated subsequently, which yields a slightly better performance as the PG baseline can stabilise the training and reduce the variance of this on-line PG method.

\paragraph{The Impact of the Different Values of $\lambda$.}
We evaluate the proposed method which largely improves the performance in our ablation studies, with more than 5\% increase on the R@1 metric of both the image and caption retrieval. The value of $\lambda$ controls the scale of the attention weights, which is with significant importance. The ablation studies show that a suitable value of $\lambda$ (20) is critical in maintaining good performance, though our method with different $\lambda$ is all with superior results.

\paragraph{The Impact of Applying a Multi-head Mechanism.}
Multi-head Mechanism is widely applied in well-known models like Transformer, often with extra improvement. We validate the positive effect of this multi-head mechanism on the proposed PG algorithm. Specifically, we apply a multi-head mechanism on the latent discrete $\mu$ and $\sigma$ values with a head number of 2. Increasing the head number would increase the computing burden, which is less practical. The empirical results reveal that the multi-head mechanism can improve the performance, by essentially reflecting different aspects of the sampled latent distribution.

\paragraph{The Impact of Utilising a Pre-trained GloVe Word Embedding.}
In the vanilla VSRN baseline, the word embedding module is trainable. We investigate the impact of a pre-trained GloVe Word Embedding module as shown in the table. Applying a pre-trained GloVe embedding can improve the matching performance slightly as it embeds some prior information.
\subsection{Visualisations}
We visualise the retrieval results and attention maps of both the image and text in Figure~\ref{img:vis1} and Figure~\ref{img:vis2}. It is clear from the figures that the attention maps can capture the expected image regions, and the language attention maps can reflect the important semantics. Some incorrect examples are also provided in the figures, which have similar semantic contents or have similar visual layouts. Visualisation on the training loss curves and the reward function curve are presented in Figure~\ref{img:curves}. The Triplet loss, Instance Loss and Text Decoding Loss all decrease as the training is performed. The reward value increases which validates the proposed Discrete-continuous PG method.

\section{Conclusions}
In this paper, we propose a novel policy gradient-based attention mechanism to transform the image and text embedding to a common space and optimise them towards higher AP. To model complex action space in the attention weights sampling, we propose a Discrete-continuous action space policy gradient algorithm, with a compound action space distribution. Comprehensive experiments on two widely-used benchmark datasets validate the effectiveness of the proposed method, leading to state-of-the-art performance.

\section{Acknowledgements}
This work is supported by the National Natural Science Foundation of China (61772524, 62002172), the Natural Science Foundation of Shanghai (20ZR1417700) and CAAI-Huawei MindSpore Open Fund.
{\small
\bibliographystyle{ieee_fullname}
\bibliography{cvpr}
}

\end{document}